\documentclass[11pt,a4paper]{article}
\PassOptionsToPackage{pdfusetitle,pdfauthor={Huiming Jin, Liwei Cai, Yihui Peng, Chen Xia, Arya D. McCarthy, Katharina Kann}}{hyperref}
\usepackage[hyperref]{acl2020}
\usepackage{times}
\usepackage{amsmath}
\usepackage{booktabs}
\usepackage{multirow}
\usepackage{arydshln}
\usepackage{subcaption}
\usepackage{graphicx}
\usepackage{bbm}
\usepackage[ruled]{algorithm2e}
\usepackage{latexsym}

\usepackage{cleveref}
\crefname{assumption}{Assumption}{Assumptions}

\usepackage{microtype}

\usepackage{adjustbox}
\usepackage{dirtytalk}
\usepackage{todonotes}
\usepackage{xcolor} 

\newcommand{\lang}[1]{\textsc{\MakeLowercase{#1}}}

\def\Snospace~{\S{}}

\aclfinalcopy
\def\aclpaperid{2072}

\DeclareMathOperator*{\argmax}{arg\,max}
\newtheorem{assumption}{Assumption}

\title{Unsupervised Morphological Paradigm Completion}

\DeclareSymbolFont{extraup}{U}{zavm}{m}{n}
\DeclareMathSymbol{\varheartsuit}{\mathalpha}{extraup}{86}
\DeclareMathSymbol{\vardiamondsuit}{\mathalpha}{extraup}{87}

\author{
Huiming Jin, Liwei Cai, Yihui Peng, Chen Xia \\
Language Technologies Institute \\
Carnegie Mellon University \\
\texttt{\{huimingj,liweicai,yihuip,chenxia\}@cs.cmu.edu} \\\AND
Arya D. McCarthy \\
Johns Hopkins University \\
\texttt{arya@jhu.edu} \\\And
Katharina Kann \\
University of Colorado \\
\texttt{katharina.kann@colorado.edu}
}

\date{}
 
\newcommand{\word}[1]{{\em #1}}

\begin{document}
\maketitle
\begin{abstract}
We propose the task
of \emph{unsupervised morphological paradigm completion}. Given only raw text and a lemma list, the task consists of generating the morphological paradigms, i.e., all inflected forms, of the lemmas. 
From a natural language processing (NLP) perspective, this is a challenging unsupervised task, and high-performing systems have the potential to improve tools for low-resource languages or to assist linguistic annotators. From a cognitive science perspective, this can shed light on how children acquire morphological knowledge. 
We further introduce a system for the task, which generates morphological paradigms via the following steps: (i) \textsc{edit tree} retrieval, (ii) additional lemma retrieval, (iii) paradigm size discovery, and (iv) inflection generation. 
We perform an evaluation on 14 typologically diverse languages.
Our system outperforms trivial baselines with ease and, for some languages, even obtains a higher accuracy than minimally supervised systems.\footnote{Our implementation is available under \url{https://github.com/cai-lw/morpho-baseline}.}
\end{abstract}

\section{Introduction}
Morphologically rich languages express syntactic and semantic properties---like \textit{tense} or \textit{case}---of words through inflection, i.e., changes to the surface forms of the words.
The set of all inflected forms of a \emph{lemma}---the canonical form---is called its \textit{paradigm}. While English does not manifest a rich inflectional morphology, Polish verbs have around a hundred different forms \citep{sadowska2012polish}, and Archi paradigms, an extreme example, can have over 1.5 million slots \cite{kibrik1977opyt}.  

\begin{figure}[t]
    \centering
    \includegraphics[width=\linewidth]{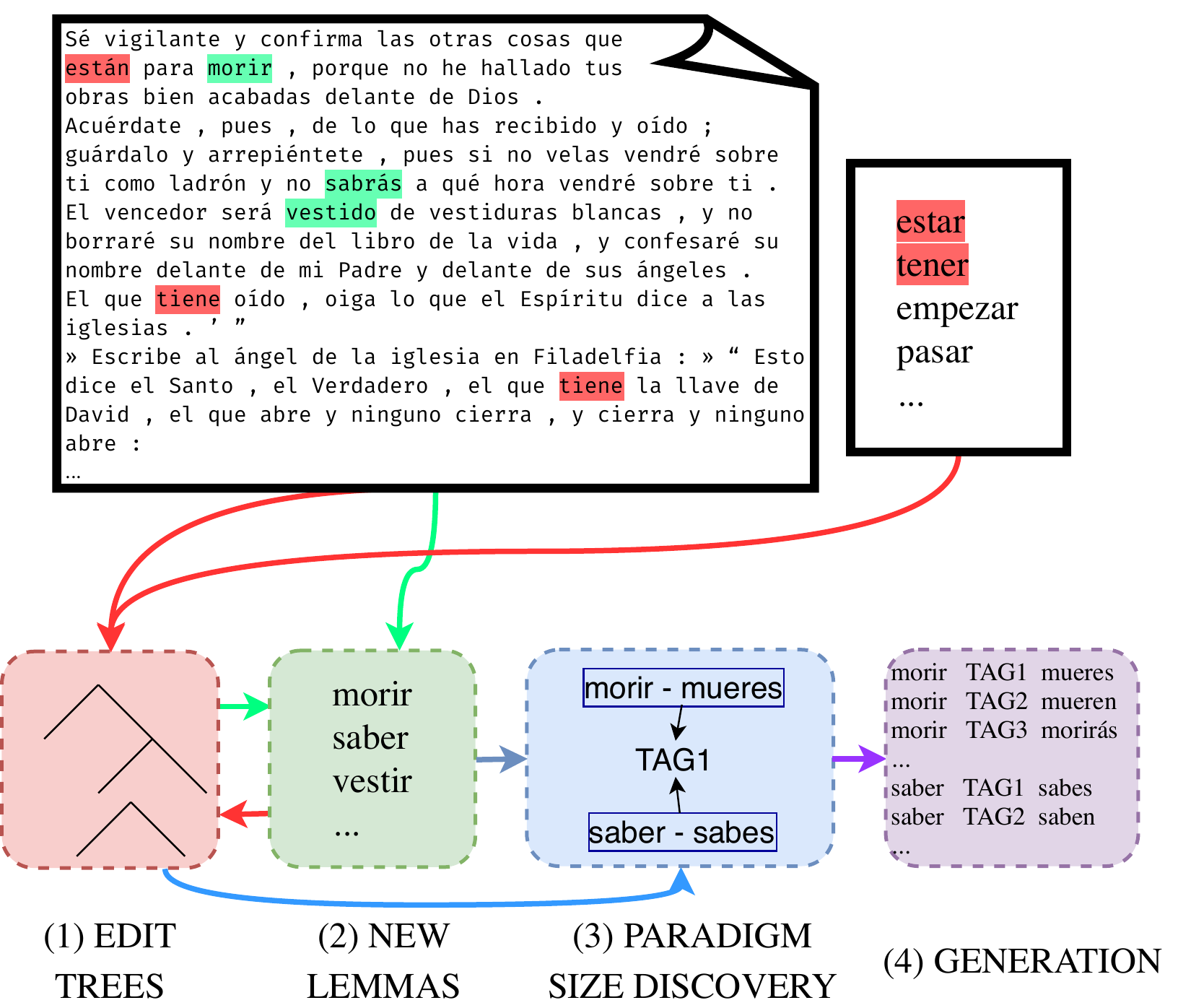}
    \caption{Our unsupervised paradigm completion system, which takes raw text and a lemma list as inputs. 
    We describe it in detail in \autoref{sec:methodology}.
    }
    \label{fig:my_label}
\end{figure}{}

Morphologically rich languages constitute a challenge for natural language processing (NLP) systems: because each lemma can take on a variety of surface forms, the frequency of each individual inflected word decreases drastically. 
This yields problems for speech recognition \cite{creutz-etal-2007-analysis}, parsing \cite{seeker-cetinoglu-2015-graph}, and keyword spotting \cite{narasimhan-etal-2014-morphological}, \textit{inter alia}.
For unsupervised machine translation, \citet{guzman-etal-2019-flores} encounter difficulties when translating into the morphologically rich languages Nepalese and Sinhalese.

Children acquire morphological knowledge from raw utterances and, in particular,
without access to explicit morphological information \cite{berko1958child}.
Do they have an innate capacity that enables them to learn a language's morphology? Or can morphology be learned in an unsupervised fashion? 
This question---in addition to practical considerations like benefits for the aforementioned NLP tasks---has motivated work on unsupervised morphological analyses \citep{goldsmith-2001-unsupervised,creutz-2003-unsupervised}.
To the best of our knowledge, no previous work has considered unsupervised morphological \textit{generation}.\footnote{\citet{kann-etal-2017-one}, which performs a zero-shot inflection experiment, uses prior information about paradigm size and related languages and, thus, cannot draw the same conclusions.} However, over the last few years, there has been a lot of progress on morphological generation tasks with limited amounts of supervision, in particular on morphological inflection \cite{cotterell-etal-2018-conll} and paradigm completion \cite{kann-schutze-2018-neural}, which can potentially be leveraged for unsupervised solutions.

Here, we fill the gap between unsupervised morphological analysis and morphological generation with limited training data by proposing the task of \textit{unsupervised morphological paradigm completion}. That is, we aim to construct and fill inflection tables
exclusively from raw text and a lemma list for a known part of speech (POS), in a situation similar to those encountered by field linguists. 
We further present a system for the task (see \autoref{fig:my_label}) which employs state-of-the-art methods common in NLP and computational morphology. 
It performs the following four steps: 
(i) \textsc{edit tree} \citep{chrupala2008towards} retrieval (\autoref{sec:form-change}), (ii) additional lemma retrieval (\autoref{sec:lemmas}), (iii) paradigm size discovery using distributional information (\autoref{sec:paradigmsize}), and (iv) inflection generation (\autoref{sec:generation}). 

To evaluate our approach, we design a metric for unsupervised paradigm completion, best-match accuracy (\autoref{sec:eval_metrics}), and experiment on 14 languages from 7 families.
As we are tackling a novel task with no baselines in the NLP literature, we perform an extensive ablation study to demonstrate the importance of all steps in our pipeline. We further show that our system outperforms trivial baselines and, for some languages, even obtains higher accuracy than a minimally supervised system.

\section{Related Work }

\paragraph{Morphological Generation }
Versions of our task with varying degrees of supervision---though never totally unsupervised---have been explored in the past.
\citet{yarowsky-wicentowski-2000-minimally} is the previous work most similar to ours. They also assume raw text and a word list as input, but additionally require knowledge of a language's consonants and vowels, as well as canonical suffixes for each part of speech. \citet{dreyer-eisner-2011-discovering} assume access to seed paradigms to discover paradigms in an empirical Bayes framework.
\citet{ahlberg-etal-2015-paradigm} and \citet{hulden-etal-2014-semi} combine information about paradigms and word frequency from corpora to perform semi-supervised paradigm completion. Our work differs from them in that we do \emph{not} assume any gold paradigms to be given. 

\citet{durrett-denero-2013-supervised}, \citet{nicolai-etal-2015-inflection}, and \citet{faruqui-etal-2016-morphological} explore a fully supervised approach, learning morphological paradigms from large annotated inflection tables. 
This framework has evolved into the SIGMORPHON shared tasks on morphological inflection  \citep{cotterell-etal-2016-sigmorphon}, which have sparked further interest in morphological generation \cite{kann-schutze-2016-single,aharoni-goldberg-2017-morphological,bergmanis-etal-2017-training,makarov-etal-2017-align,zhou-neubig-2017-multi,kann-schutze-2018-neural}. 
We integrate two systems \cite{cotterell-etal-2017-conll,makarov-clematide-2018-neural} produced for SIGMORPHON shared tasks into our framework for unsupervised morphological paradigm completion.

\paragraph{Morphological Analysis }
Most research on unsupervised systems for morphology aims at developing approaches to segment words into their smallest meaning-bearing units, called \textit{morphemes} \cite{goldsmith-2001-unsupervised,creutz-2003-unsupervised,creutz2007unsupervised,snyder-barzilay-2008-unsupervised}. Unsupervised morphological paradigm completion differs from segmentation in that, besides capturing how morphology is reflected in the word form, it also requires correctly clustering transformations into paradigm slots as well as generating unobserved forms. 
The model by \newcite{xu-etal-2018-unsupervised-morphology} recovers something akin to morphological paradigms. However, those paradigms are a means to a segmentation end, and \newcite{xu-etal-2018-unsupervised-morphology} do not explicitly model information about the paradigm size as required for our task.

Other unsupervised approaches to learning morphological analysis and generation rely on projections between word embeddings \citep{soricut-och-2015-unsupervised, narasimhan-etal-2015-unsupervised}; however, these approaches rely on billions of words to train embeddings; at a minimum, \citet{narasimhan-etal-2015-unsupervised} use 129 million word tokens of English Wikipedia. As we will describe later on (\autoref{subsec:data}), we, in contrast, are concerned with the setting with mere thousands of sentences.

For a detailed survey of unsupervised approaches to problems in morphology, we refer the reader to \citet{hammarstrom-borin-2011-unsupervised}.

\paragraph{SIGMORPHON 2020: Unsupervised Morphological Paradigm Completion} After multiple shared tasks on morphological inflection starting with \newcite{cotterell-etal-2016-sigmorphon}, in 2020, SIGMORPHON (the ACL special interest group on computational morphology and phonology) is organizing its first shared task on unsupervised morphological paradigm completion.\footnote{\url{https://sigmorphon.github.io/sharedtasks/2020/task2/}} The system presented here is the official shared task baseline system. The first other approach
applicable to this shared task 
has been developed by \newcite{erdmann2020}. Their  pipeline system is similar in spirit to ours, but the individual components are different, e.g., a transformer model \cite{vaswani2017attention} is used for inflection generation.

\section{Formal Task Description}
\label{sec:formal-description}
Given a corpus $\mathcal{D}=w_1,\dots,w_{|\mathcal{D}|}$ with a vocabulary $V$ of word types $\{w_i\}$ and a lexicon $\mathcal{L} = \{\ell_j\}$ with $|\mathcal{L}|$~lemmas belonging to the same part of speech, 
the task of unsupervised morphological paradigm completion consists of
generating the paradigms~$\{\pi(\ell)\}_{\ell\in\mathcal{L}}$ of the entries in the lexicon.

Following \citet{matthews1972inflectional} and \citet{aronoff1976word}, we treat a paradigm as a \emph{vector of inflected forms} belonging to a lemma~$\ell$.
Paradigm completion consists of predicting missing slots in the paradigm $\pi(\ell)$:
\begin{equation}
    \pi(\ell) = \left\langle f(\ell, \vec{t}_\gamma)\right\rangle_{\gamma \in \Gamma(\ell)}\text{,}
\end{equation}
where $f : \Sigma^* \times \mathcal{T} \to \Sigma^*$ transforms a lemma into an inflected form,\footnote{We assume a discrete alphabet of symbols $\Sigma$.
} $\vec{t}_\gamma \in \mathcal{T}$ is a vector of inflectional features describing paradigm slot $\gamma$,
and $\Gamma(\ell)$ is the set of slots in lemma $\ell$'s paradigm. Since we only consider lemmas that belong to the same part of speech, we will use $\Gamma$ and $\Gamma(\ell)$ interchangeably in the following. Furthermore, we will denote $f(\ell, \vec{t}_\gamma)$ as $f_{\gamma}(\ell)$ for simplicity.

\paragraph{Remarks on Task Design } In general, not all paradigm entries will be present in the corpus $\mathcal{D}$. Thus, the task requires more than a keyword search.

On another note, it is not necessary to predict the features $\vec{t}_\gamma$ corresponding to each slot~$\gamma$; as the exact denotation of features is up to human annotators, they cannot be inferred by unsupervised systems. For our task, it is enough to predict the ordered vector $\pi(\ell)$ of inflected forms. 

\section{Methodology}
\label{sec:methodology}
Our system implicitly solves 
two subtasks:
(i) determining the number of paradigm slots; and (ii) generating the inflected form corresponding to each paradigm slot for each lemma.
It is organized as a pipeline consisting of multiple components. 
Our system is highly modular: individual components
can be exchanged easily.
In the remainder of this section, we will dedicate each subsection to one component.

\subsection{Retrieval of Relevant \textsc{Edit Trees}}
\label{sec:form-change}
The first component in our pipeline identifies words in the corpus \(\mathcal{D}\) which could belong to the paradigm of one of the lemmas in the lexicon \(\mathcal{L}\). We call those words \textit{paradigm candidates}.
It then uses the discovered paradigm candidates to identify \textsc{edit tree} \citep{chrupala2008towards}
operations that correspond to valid inflections.

\paragraph{Paradigm Candidate Discovery }
For most paradigms, all participating inflected forms share some characteristics---usually substrings---which humans use to identify the paradigm any given word belongs to.
Given a pair~$(\ell, w)$ of a lemma~$\ell$ and a word form~$w$, 
the first step in our pipeline is to determine whether $w$ is a paradigm candidate for lemma $\ell$.
For example, \textit{studied} is likely to be an inflected form of the English lemma \textit{study}, 
while \textit{monkey} is not. 
We identify paradigm candidates $C_{\ell}$ of a lemma $\ell$ by computing the longest common substring (LCS) between $\ell$ and $w$ for all words $w$ in the vocabulary $V$. If the ratio between the LCS's length and the length of~$\ell$ is higher than a threshold $\lambda_P$, $w$ is a paradigm candidate for~$\ell$:
\begin{align}
    C_{\ell} = \left\{w\in V\left|\frac{|\mathrm{LCS}(\ell, w)|}{|\ell|} > \lambda_P\right.\right\}.
\end{align}

\paragraph{\textsc{Edit tree} Discovery } Surface form changes, which we denote as $\psi$, define a modification of a word's surface form. 
Our system employs \textsc{edit trees} \cite{chrupala2008towards} to represent $\psi$.

\begin{figure}[tbp] 
    \centering
    \includegraphics[width=0.75\linewidth]{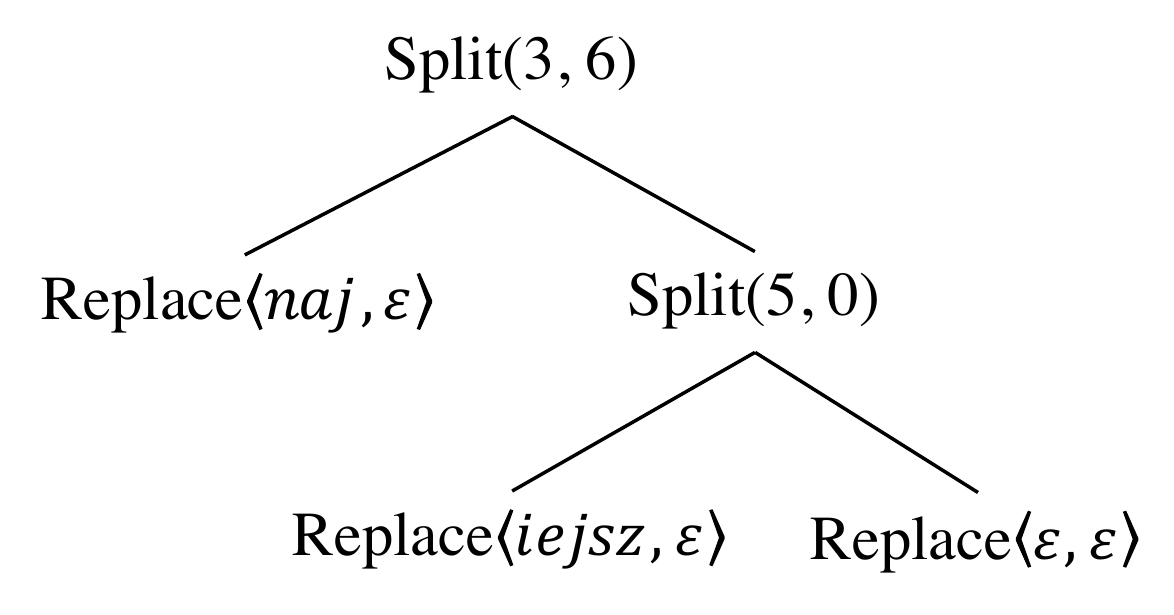}
    \caption{Visualization of the \textsc{edit tree} constructed from {\it najtrudniejszy} to {\it trudny} \cite{chrupala2008towards}.}
    \label{fig:et} 
\end{figure} 

Given two strings $x_{[1\dots n]}$ and $x_{[1\dots m]}'$, $\textsc{edit tree}(x_{[1\dots n]}, x_{[1\dots m]}')$  
is constructed~by first determining the LCS between $x$ and $x'$. 
We then recursively model the substrings before and after the LCS. If the length of the LCS is zero, the \textsc{edit tree} consists of the substitution operation of the first string with the second.
For example, $\textsc{edit tree}(\textit{najtrudniejszy}, \textit{trudny})$\footnote{Polish form--lemma pair meaning {\it hardest} and {\it hard}. Example from \newcite{chrupala2008towards}.} could be visualized as in \autoref{fig:et}, where $\mathrm{Split}(i, j)$ represents taking the substring $x_{[i\dots n-j]}$.

An \textsc{edit tree} can be applied to new input strings. The \textsc{edit tree} in \autoref{fig:et}, for example, could be applied to \emph{naj}\textbf{\emph{apple}}\emph{iejsz}\textbf{\emph{s}}, and the resulting string would be \textbf{\emph{apples}}.
Note that not all
\textsc{edit trees} can be applied to all strings. For example, the \textsc{edit tree} in \autoref{fig:et} can only be applied to words starting with \textit{naj}.

Our system constructs \textsc{edit trees} from all pairs $(\ell, w)$ of lemmas $\ell$ and their paradigm candidates $w$ and counts their frequencies:
\begin{align}
    n_{\psi} = \sum_{\substack{\ell\in \mathcal{L}\\w\in C_{\ell}}}\mathbbm{1}\left[\textsc{edit tree}(\ell, w) = \psi\right]\text{.}
\end{align}
It then discards \textsc{edit trees} with frequencies $n_{\psi}$ below a threshold $\lambda_{FC}(|\mathcal{L}|)$, which is a function of the size of the lexicon $\mathcal{L}$:
\begin{align}
    \lambda_{FC}(|\mathcal{L}|) = \max\left\{2, \phi_{FC}\cdot|\mathcal{L}|\right\}, \label{eq:lambda_fc}
\end{align}
where $\phi_{FC}\in \mathbbm{R}$ is a hyperparameter. The idea is that an \textsc{edit tree} is only valid if it can be applied to multiple given lemmas. \textsc{edit trees} which we observe only once are always considered unreliable. 
Our system then retains a set of frequent surface form changes 
\begin{align}
    \Psi = \left\{\psi \mid n_{\psi}\geq\lambda_{FC}(|\mathcal{L}|)\right\}
\end{align}
represented by \textsc{edit trees}. Assuming an one-to-one mapping between surface form changes and paradigm slots (that is, that \(|\Psi| = |\Gamma|\) and that each \(\psi\) is equivalent to a particular inflection function $f_\gamma$), we now have a first basic paradigm completion system (\textbf{\textsc{PCS-I}}), which operates by applying all suitable \textsc{edit trees} to all lemmas in our lexicon.

\paragraph{Complexity and Runtime}
The time complexity to compute
$\textsc{edit tree}(x_{[1\dots n]}, x_{[1\dots m]}')$
is $O(\max^3\{n, m\})$.
Computing \textsc{edit trees} can trivially be parallelized, and in practice this computation does not take much time.

\subsection{Retrieval of Additional Lemmas}
\label{sec:lemmas}
Since we assume a low-resource setting, our lexicon is small ($\leq 100$ entries). However, the more lemmas we have, the more confident we can be that the \textsc{edit trees} retrieved by the first component of our system represent valid inflections.
An intuitive method to obtain additional lemmas would be to train a lemmatizer and to generate new lemmas from words in our corpus. 
However, due to the limited size of our initial lemma list, such a lemmatizer would most likely not be reliable.

The second component of our system employs another method,
which guarantees that additionally retrieved lemmas are valid words: 
It is based on the intuition that a word $w \in V$ is likely to be a lemma if the pseudo--inflected forms of $w$, obtained by applying the \textsc{edit trees} from \autoref{sec:form-change}, also appear in $V$. 
For a word $w \in V$, we say it is a discovered lemma if $w \notin \mathcal{L}$ and
\begin{align}
    \sum_{\psi \in \Psi}\mathbbm{1}\left[\psi(w) \in V\right] &> \lambda_{\mathit{NL}}(|\Psi|)
\end{align}
for
\begin{align}
\lambda_{\mathit{NL}}(|\Psi|) = \max\left\{3, \phi_{\mathit{NL}}\cdot|\Psi|\right\},
\end{align}
with $\phi_{\mathit{NL}}\in \mathbbm{R}$ being a hyperparameter. Similar to \autoref{eq:lambda_fc}, $\lambda_{\mathit{NL}}$ depends on the number of discovered \textsc{edit trees}, but is never smaller than $3$. We set this minimum to require evidence for at least two transformations in addition to the identity.

We can now bootstrap by iteratively computing additional paradigm candidates and \textsc{edit trees}, and then retrieving more lemmas. We denote the paradigm completion systems resulting from one and two such iterations as \textbf{\textsc{PCS-II-a}} and \textbf{\textsc{PCS-II-b}}, respectively.
Since we cannot be fully confident about retrieved lemmas, we associate each additional lemma with a weight
$\theta_{\ell} = \theta_{\mathit{NL}}^{it}$,
where $\theta_{\mathit{NL}}$ is a preset hyperparameter and $it$ identifies the iteration in which a lemma is added, i.e., $it=0$ for gold lemmas and $it=i$ for  lemmas retrieved in the $i$th iteration.
The weights $\theta_{\ell}$ are used in later components of our system. 

\subsection{Paradigm Size Discovery}
\label{sec:paradigmsize}
\begin{figure}[tbp]
    \begin{subfigure}{0.49\linewidth}
        \includegraphics[width=\linewidth]{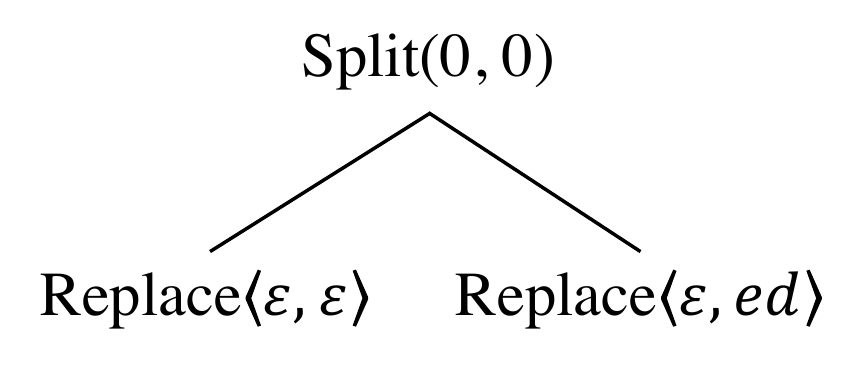}
        \caption{} \label{fig:et-work}
    \end{subfigure}
    \hspace*{\fill}
    \begin{subfigure}{0.49\linewidth}
        \includegraphics[width=\linewidth]{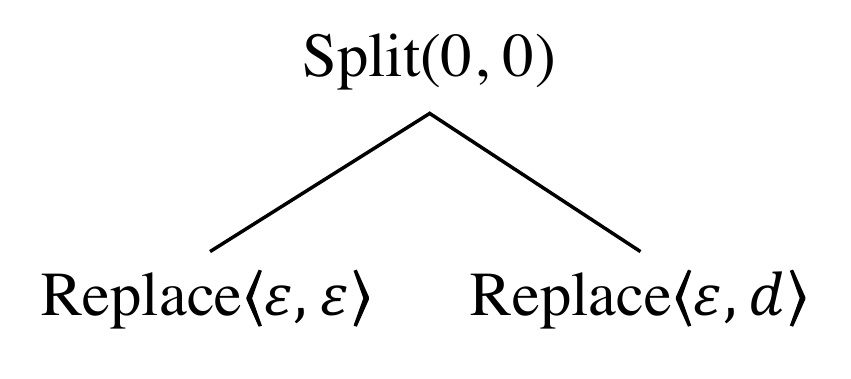}
        \caption{} \label{fig:et-continue}
    \end{subfigure}
    \caption{Visualization of the \textsc{edit trees} representing (a) \textit{work} $\mapsto$ \textit{worked} and (b) \textit{continue} $\mapsto$ \textit{continued}.} \label{fig:et-work-continue}
\end{figure}

Until now, we have assumed a one-to-one mapping between paradigm slots and surface form changes. However, different \textsc{edit trees} may indeed represent the same inflection. For example, the past tense inflection of verbs in English involves 
multiple \textsc{edit trees}, as shown in \autoref{fig:et-work-continue}.

Thus, the next step is to group surface form changes based on the paradigm slots they realize.

\subsubsection{One \textsc{Edit Tree} per Lemma and Slot}
Our algorithm for grouping surface form changes is based on two assumptions. 
First, since
\textsc{edit trees} are extracted from ($\ell$, $w$) pairs, different \textsc{edit trees} belonging to the same paradigm slot cannot be extracted from the same lemma.\footnote{Exceptions to this do exist. However, they are rare enough that we do not expect them to hurt our algorithm.} 
Thus:

\begin{assumption}
\label{thm:onetreeperlemmaslot}
For each lemma, at most one inflected form 
per paradigm slot 
can be found in the corpus.
\end{assumption}

Formally, for a multi-to-one mapping from \textsc{edit trees} to paradigm slots $z: \Psi \to \Gamma$, we define the \textsc{edit tree} set \(\Psi_{\gamma}\) of a potential paradigm slot \(\gamma\) as $\Psi_{\gamma} = \left\{\psi \mid z(\psi)=\gamma\right\}$, with $\bigcup_{\gamma} \Psi_{\gamma} = \Psi$. Then, for any lemma $\ell \in \mathcal{L}$ and proposed paradigm slot $\gamma \in \Gamma$, 
we have:
\begin{align}
    \left|\left\{w\in V \mid \psi(\ell)=w \wedge \psi \in \Psi_{\gamma}\right\}\right| \leq 1.
\end{align}

\subsubsection{One Paradigm Slot per \textsc{Edit Tree}}
Our second assumption is a simplification,\footnote{This assumption ignores syncretism.} but helps to reduce the search space during clustering: 

\begin{assumption}
\label{thm:oneslotpertree}
Each surface form change $\psi \in \Psi$ belongs to exactly one paradigm slot.
\end{assumption}

Formally, we partition \(\Psi\) into disjoint subsets:
\begin{align}
    \Psi_{\gamma} \cap \Psi_{\gamma'} = \emptyset\qquad \forall \gamma \neq \gamma'.
\end{align}

\subsubsection{Paradigm Slot Features and Similarity} 
In addition to \Cref{thm:onetreeperlemmaslot,thm:oneslotpertree}, we make use of a feature function $r(\gamma)$ and a score function $s(r(\gamma), r(\gamma'))$, which measures the similarity between two potential paradigm slots.

\begin{figure}[tbp]
    \centering
    \includegraphics[width=\linewidth]{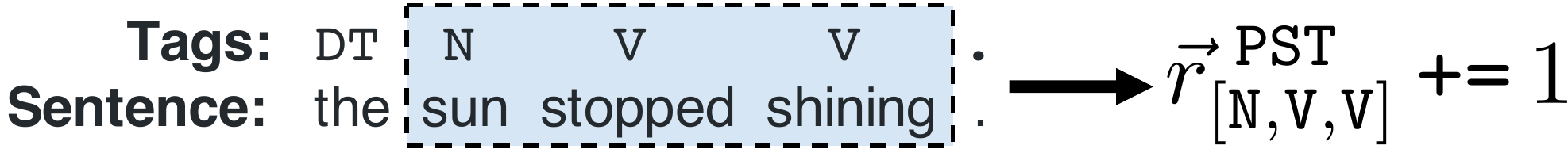}
    \caption{An example of the distributionally informed feature function with window size $3$ for the past tense slot (\texttt{PST}). \textit{stop} $\in \mathcal{L}$ and $f_{\texttt{PST}}(stop) = stopped$.
    When the sliding window arrives to this instance of \textit{stopped}, $\vec{r}_{[\texttt{N}, \texttt{V}, \texttt{V}]}^{\,\,\texttt{PST}}$ is increased by~$1$.\protect\footnotemark}
    \label{fig:feature_function}
\end{figure}{}
\footnotetext{Note that this example is simplified. The system does not need to actually know the tags like \texttt{PST} and \texttt{V}.}

Our \emph{feature function} 
makes the connection between paradigm slots and 
the instances of inflected forms in the corpus by utilizing the part-of-speech (POS) tags as context information. In our implementation, we employ an unsupervised POS tagger \cite{stratos-etal-2016-unsupervised} to extract 
tags for each word in the corpus $\mathcal{D}$.

This tagger
assigns an \textit{anchor word}, i.e., a pseudo POS tag, to each word $w_i$ in the corpus by using an \textbf{anchor hidden Markov model (HMM)} with $8$ hidden states. Our feature function counts the tag tuples within a sliding window centered on each instance of inflected forms of a potential paradigm slot.

Formally, we denote the set of lemmas that surface form change $\psi$ is applied to as $\mathcal{L}_{\psi}$, the set of lemmas that express a potential paradigm slot $\gamma$ as $\mathcal{L}_{\gamma}$, and the corresponding inflected forms as $V_{\gamma}$:
\begin{align}
    \mathcal{L}_{\psi} & = \left\{\ell \in \mathcal{L} | \psi(\ell)=w \wedge w\in V \right\} \\
    \mathcal{L}_{\gamma} &= \bigcup_{\psi\in\Psi_{\gamma}} \mathcal{L}_{\psi} \\
    V_{\gamma} &= \bigcup_{\psi\in\Psi_{\gamma}}\left\{\psi(\ell)\right\}_{\ell\in\mathcal{L}_{\psi}}.
\end{align}
We further refer to the set of available POS tag labels as $P=\left\{p_1, \dots, p_{8}\right\}$. For a corpus $\mathcal{D} = w_1, \dots, w_{|\mathcal{D}|}$, a window size $2d+1$, and a potential paradigm slot $\gamma$, our feature function is defined as:
\begin{align}
    \vec{r}^{\,\,\gamma} &= r(\gamma; \mathcal{D}, 2d+1),
\end{align}
where $\vec{r}^{\,\,\gamma}$ is a vector with one dimension for each possible tag tuple sequence of length 2d+1. Its values are computed as:
\begin{align}
   \vec{r}^{\,\,\gamma}_{\left[p_{j_1},\dots,p_{j_{2d+1}}\right]} &= \sum_{w_i\in \mathcal{D}}\mathbbm{1}\big[w_i\in V_{\gamma}\\ \wedge\,\hat{p}_{i-d}&=p_{j_1}\wedge\dots\wedge \hat{p}_{i+d}=p_{j_{2d+1}}\big] \nonumber\\
 \forall&\left(\hat{p}_{j_1},\dots,\hat{p}_{j_{2d+1}}\right)\in P^{2d+1}. \nonumber
\end{align}

In practice, we initialize $\vec{r}^{\,\,\gamma}$ to the zero vector, and iterate the sliding window over the corpus. When the central word $w_i \in V_\gamma$, 
the corresponding value of $\vec{r}^{\,\,\gamma}$ at the POS tuple within the sliding window is incremented by $1$. \autoref{fig:feature_function} shows an example in English.

We assume that paradigm slots $\gamma$ and $\gamma'$ are similar if the words in $V_{\gamma}$ and $V_{\gamma'}$ frequently appear in similar contexts, i.e., within similar tag sequences. 
With the feature function defined above, our system uses \textit{cosine similarity} as the score function~$s$.

\begin{algorithm}[tb]
\small
\SetAlgoLined
\KwResult{$\Gamma, \{\Psi_{\gamma}\}_{\gamma\in\Gamma}$}
 Initialize $\Gamma$ s.t. $|\Gamma| = |\Psi|$ and $f_{\gamma_i} = \psi_i$ for all $\psi_i \in \Psi$\;
 Initialize $\Psi_{\gamma_i} = \{\psi_i\}$ for all $\psi_i \in \Psi$\;
 \While{$\exists\,(\gamma,\gamma') \quad\mathrm{s.t.}\quad \mathcal{L}_{\gamma}\cap \mathcal{L}_{\gamma'}=\emptyset$}{
  $\overrightarrow{score}\gets \vec{0}$\;
  \For{$(\gamma,\gamma')\quad s.t.\quad \mathcal{L}_{\gamma}\cap \mathcal{L}_{\gamma'}=\emptyset$}{
    $\overrightarrow{score}_{(\gamma,\gamma')}\gets s(r(\gamma), r(\gamma'))$\;
  }
  $(\hat{\gamma},\hat{\gamma}')=\argmax_{(\gamma,\gamma')}\{\overrightarrow{score}_{(\gamma,\gamma')}\}$\;
  \eIf{$\overrightarrow{score}_{(\hat{\gamma},\hat{\gamma}')} > \lambda_{S}$}{
   denote the new paradigm slot $\gamma_{\text{merge}}$\;
   $\Psi_{\gamma_{\text{merge}}} = \Psi_{\hat{\gamma}}\cup\Psi_{\hat{\gamma}'}$\;
   $\Gamma\gets(\Gamma\setminus\{\hat{\gamma},\hat{\gamma}'\})\cup\{\gamma_{\text{merge}}\}$\;
   }{
   \textbf{break}\;
  }
 }
 \caption{Surface Form Change Grouping}
 \label{alg:grouping}
\end{algorithm}

We then develop \autoref{alg:grouping} to group one-to-one mappings from surface form changes to paradigm slots into many-to-one mappings. The idea is to iteratively merge the most similar slots if this is not violating \cref{thm:onetreeperlemmaslot} until the similarity gets too low. $\lambda_{S} \in (0, 1)$ is a threshold parameter.\footnote{This does not result in a functional paradigm completion system, since we still lack a method to decide which surface form change to apply to realize a paradigm slot for a lemma.}

\subsection{Generation}
\label{sec:generation}
Now, one paradigm slot can be represented by multiple \textsc{edit trees}.
Our system, thus, needs to learn to apply the correct transformation for a combination of lemma and paradigm slot.
However, mapping lemmas and paradigm slots to inflected forms corresponds exactly to the morphological inflection task, which has been the subject of multiple shared tasks over the last years \cite{cotterell-etal-2018-conll}.

Our morphological inflection models take $\left(\text{slot}, \text{lemma}, \text{word}\right)$ tuples extracted by the previous components of our system as training data. Formally, they are trained on the training set:
\begin{align}
    \left\{(\vec{t}_\gamma, \ell, f_\gamma(\ell)) \mid \gamma\in \Gamma \wedge \ell\in\mathcal{L}_{\gamma} \wedge f_\gamma(\ell) \in V\right\}.
\end{align}

We explore two morphological inflection models from the literature.

\paragraph{Affix Editing} The baseline system of the  CoNLL--SIGMORPHON 2017 shared task\footnote{\url{https://github.com/sigmorphon/conll2017/tree/master/baseline}} \cite{cotterell-etal-2017-conll} is a simple approach, which is very suitable for low-resource settings. The system breaks
each word into \textsc{prefix}, \textsc{stem}, and \textsc{suffix}, and then stores the \textsc{prefix} editing rules and the \textsc{stem}+\textsc{suffix} editing rules. At test time, it applies the longest possible \textsc{prefix} and \textsc{stem}+\textsc{suffix} editing rules to the input lemma. We denote the surface form change grouping in combination with this system as \textbf{PCS-III-C}.
 
\paragraph{Transducer-Based Hard-Attention} We further experiment with a transducer-based hard-attention model \cite{makarov-clematide-2018-imitation}. Unlike widely used soft-attention sequence-to-sequence models \cite{bahdanau2015neural}, which predict the target tokens directly, it predicts edit action sequences to transform the input sequence into outputs, and it disposes of a hard attention mechanism. 
We denote the surface form change grouping in combination with this system as \textbf{PCS-III-H}.

\section{Experiments}
\subsection{Data}
\label{subsec:data}
To evaluate our approach in a real-world setting, 
we restrict our data to resources typically available to a field linguist: a small written corpus ($\leq$ 100k tokens) and a small lexicon.

For our corpora, we use the JHU Bible Corpus \cite{mccarthy-etal-2020-johns}, which allows future work to build systems in 1600 languages. The Bible is frequently available even in low-resource languages: Ethnologue identifies 3,995 written languages, and the New Testament has been translated into 2,246. The Bible is also highly representative of a language's \textit{core vocabulary}: \citet{Resnik1999} find high overlap with both the Longman Dictionary of Contemporary English \cite{summers1995longman} and the Brown Corpus \cite{francis1964brown}. Furthermore, the Bible is multi-parallel and, thus, allows for a fair comparison across languages without confounds like domain.

For \textit{evaluation} of our methods only, we additionally obtain ground-truth morphological paradigms from UniMorph \citep{kirov-etal-2018-unimorph}, which provides paradigms for over 100 languages. 

From the intersection of languages in the Bible and UniMorph, we select 14 typologically diverse languages from 7 families, each of which display inflectional morphology:
Basque (\lang{EUS}), Bulgarian (\lang{BUL}), English (\lang{ENG}), Finnish (\lang{FIN}), German (\lang{DEU}), Kannada (\lang{KAN}), Navajo (\lang{NAV}), Spanish (\lang{SPA}), and Turkish (\lang{TUR}) as test languages, and Maltese (\lang{MLT}), Persian (\lang{FAS}), Portuguese (\lang{POR}), Russian (\lang{RUS}), and Swedish (\lang{SWE}) for development.
To create test data for all and development data for our development languages, we sample 100 paradigms for each set
from UniMorph, then take their lemmas as our lexicon~\(\mathcal{L}\).\footnote{For Basque, Kannada, and Maltese, we only take 20 paradigms for each set, due to limited availability.}

\subsection{Baselines and Skylines}

\paragraph{Lemma Baseline (LB) } 
Our first, trivial baseline predicts inflected forms identical to the lemma for all paradigm slots.
We compare to one version of this baseline that has access to the ground-truth paradigm size  (\textbf{LB-Truth}), 
and a second version 
which predicts the paradigm size as the average over the development languages
(\textbf{LB-Dev}). 

\paragraph{One/Ten-Shot Inflection Model } Our second baseline could be seen as a \textit{skyline}, since it leverages morphological information our proposed system does not have access to.
In particular, we train the baseline system of CoNLL--SIGMORPHON 2017 \cite{cotterell-etal-2017-conll} on one (\textbf{CoNLL17-1}) and ten (\textbf{CoNLL17-10}) paradigms. For this, we randomly sample paradigms from UniMorph, excluding those in our test data.

\subsection{Hyperparameters}
We choose the hyperparameters by grid search over intuitively reasonable ranges, using the development languages. No test language data is seen before final testing. Note also that only the corpus and the lexicon can be accessed by our system, 
and no ground-truth morphological information (including paradigm size) is given. 

Our final hyperparameters are $\lambda_P=0.5$, $\phi_{FC} = 0.05$, $\phi_{\mathit{NL}} = 0.2$, $\theta_{\mathit{NL}}=0.5$, and $\lambda_{S} = 0.3$. The window size $2d+1$ for feature extraction is set to $3$.
For \textbf{CoNLL17-1}, we average over the results for 10 different sampled paradigms for each language.
 
\subsection{Evaluation Metrics}
\label{sec:eval_metrics}

\begin{table*}
    \centering
    \begin{adjustbox}{max width=\linewidth}
    \begin{tabular}{
        l 
        c 
        c c c c c  
        c c 
    }
    \toprule
        &
        \multicolumn{7}{c}{\textbf{Test}} \\
    \cmidrule(lr){2-9}
        \textbf{Method} &
        \textbf{Test Ave.} & 
        \textbf{\lang{eus}} & 
        \textbf{\lang{bul}} &
        \textbf{\lang{eng}} &
        \textbf{\lang{fin}} &
        \textbf{\lang{deu}} &
        \textbf{\lang{kan}} &
        \textbf{\lang{nav}}
        \\
    \midrule
        LB-Truth & 4.94 / 5.25 & 0.03 / 0.05 (1659) & 1.89 / 2.16 (56) & 20.40 / 20.40 (5) & 0.96 / 0.96 (141) & 15.66 / 15.68 (29) & 1.18 / 1.19 (85) & \textbf{2.89} / \textbf{5.39} (50) \\
        LB-Dev &2.22 / 2.53 &  0.03 / 0.05 (48) & 1.89 / 2.16 (48) & 2.13 / 2.15 (48) & 0.96 / 0.96 (48) & 9.46 / 9.47 (48) & 1.18 / 1.19 (48) & \textbf{2.89} / \textbf{5.39} (48) \\
        CoNLL17-1 & 18.70 / 18.70 & 0.00 / 0.00 (1659) & 12.34 / 12.24 (56) & 59.92 / 59.92 (5) & 3.52 / 3.52 (141) & 26.71 / 26.73 (29) & 5.74 / 5.82 (85) & 0.00 / 0.00 (50) \\
        CoNLL17-10 & \textbf{35.58} / \textbf{35.56} & 0.00 / 0.00 (1659) & {\bf 43.75} / {\bf 43.58} (56) & 70.58 / 70.58 (5) & {\bf 24.51} / {\bf 24.51} (141) & \textbf{35.75} / \textbf{35.77} (29) & \textbf{34.53} / \textbf{34.51} (85) & 0.00 / 0.00 (50) \\
    \midrule
        PCS-I & 16.09 / 16.39 & \textbf{\underline{0.11}} / \textbf{\underline{0.11}} (51) & 18.01 / 19.18 (39) & 56.17 / 56.17 (6) & 3.40 / 3.40 (23) & 18.28 / 18.30 (19) & 14.29 /14.15 (209) & 2.19 / 3.74 (8) \\
        \textsc{PCS-I+II-a} & 17.60 / 17.89 & \textbf{\underline{0.11}} / \textbf{\underline{0.11}} (51) & \underline{22.72} / \underline{23.87} (52) & 56.17 / 56.17 (6) & \underline{5.70} / \underline{5.70} (71) & 20.57 / 20.56 (32) & 12.60 / 12.48 (237) & 2.19 / 3.74 (19) \\
        \textsc{PCS-I+II-b} & 17.60 / 17.89 & \textbf{\underline{0.11}} / \textbf{\underline{0.11}} (51) & \underline{22.72} / \underline{23.87} (57) & 56.17 / 56.17 (6) & \underline{5.70} / \underline{5.70} (71) & 20.57 / 20.56 (32) & 12.60 / 12.48 (237) & 2.19 / 3.74 (18) \\
        PCS-I+III-C & 15.90 / 16.19 & 0.07 / 0.10 (27) & 16.18 / 17.24 (16) & 63.20 / 63.20 (5) & 3.58 / 3.58 (15) & 5.19 / 5.18 (14) & 17.82 / 17.82 (162) & 2.19 / 3.74 (8) \\
        PCS-I+III-H & 17.45 / 17.67 & 0.04 / 0.04 (27) & 16.95 / 18.16 (16) & 62.20 / 62.20 (5) & 3.78 / 3.78 (15) & 19.98 / 19.96 (14) & 15.83 / 15.85 (162) & \underline{2.55} / \underline{4.78} (8) \\
        PCS-I+II+III-C & \underline{19.10} / \underline{19.41} &  0.07 / 0.10 (27) & 16.77 / 17.94 (16) & 72.80 / 72.80 (4) & 3.51 / 3.51 (15) & \underline{23.28} / \underline{23.31} (13) & \underline{18.03} / \underline{17.90} (161) & 2.16 / 3.81 (7) \\
        PCS-I+II+III-H & 18.76 / 19.06 & 0.06 / 0.08 (27) & 17.39 / 18.65 (16) & \textbf{\underline{74.00}} / \textbf{\underline{74.00}} (4) & 3.77 / 3.77 (15) & 18.87 / 18.89 (13) & 17.90 / \underline{17.90} (161) & 2.0 / 3.41 (7) \\
    \bottomrule
    \end{tabular}
    \end{adjustbox}
    \newline
    \vspace{0.5em}
    \newline
    \begin{adjustbox}{max width=\linewidth}
    \begin{tabular}{
        l 
        c c  
        c 
        c c c c c  
    }
    \toprule
        &
        \multicolumn{2}{c}{\textbf{Test}} & 
        \multicolumn{5}{c}{\textbf{Development}} \\
    \cmidrule(lr){2-3}
    \cmidrule(lr){4-9}
        \textbf{Method} &
        \textbf{\lang{spa}} &
        \textbf{\lang{tur}} &
        \textbf{Dev. Ave.}  &
        \textbf{\lang{mlt}} &
        \textbf{\lang{fas}} &
        \textbf{\lang{por}} &
        \textbf{\lang{rus}} &
        \textbf{\lang{swe}}
        \\
    \midrule
        LB-Truth & 1.43 / 1.43 (70) & 0.00 / 0.00 (120) & 7.94 / 7.87 & 9.21 / 7.69 (24) & 2.04 / 2.04 (136) & 6.53 / 6.53 (76) & 5.47 / 5.17 (55) & 16.44 / 17.91 (11) \\
        LB-Dev & 1.43 / 1.43 (48) & 0.00 / 0.00 (48) & 4.33 / 4.18 & 4.25 / 3.55 (52) & 2.04 / 2.04 (33) & 6.53 / 6.53 (43) & 5.47 / 5.17 (47) & 3.35 / 3.65 (54) \\
        CoNLL17-1 & 60.07 / 60.07 (70) & 0.00 / 0.00 (120) & 33.68 / 34.73 & 11.82 / 12.08 (24) & 46.68 / 46.68 (136) & 82.87 / 82.87 (76) & 7.75 / 10.98 (55) & 19.30 / 21.03 (11) \\
        CoNLL17-10 & \textbf{76.23} / \textbf{76.23} (70) & \textbf{34.88} / \textbf{34.88} (120) & \textbf{49.3} / \textbf{53.02} & 13.07 / \textbf{13.33} (24) & \textbf{64.26} / \textbf{64.26} (136) & \textbf{89.54} / \textbf{89.54} (76) & \textbf{19.22} / \textbf{36.75} (55) & \textbf{60.41} / \textbf{61.22} (11) \\
    \midrule
        PCS-I & 25.04 / 25.04 (30) & 7.40 / 7.40 (186) & 18.72 / 21.66 & 12.21 / \underline{12.00} (25) & 6.36 / 6.36 (35) & 36.47 / 36.47 (45) & 8.24 / 20.98 (22) & 30.31 / 32.47 (20) \\
        \textsc{PCS-I+II-a} & \underline{31.03} / \underline{31.03} (39) & 7.33 / 7.33 (237) & 17.06 / 20.77 & 8.45 / 8.50 (38) & 6.50 / 6.50 (52) & \underline{37.14} / \underline{37.14} (92) & 12.56 / 30.66 (53) & 20.65 / 21.06 (33) \\
        \textsc{PCS-I+II-b} & \underline{31.03} / \underline{31.03} (39) & 7.33 / 7.33 (237) & 17.02 / 20.80 & 7.83 / 7.88 (41) & \underline{6.51} / \underline{6.51} (52) & \underline{37.14} / \underline{37.14} (92) & 12.95 / \underline{31.42} (55) & 20.65 / 21.06 (33) \\
        PCS-I+III-C & 24.10 / 24.10 (29) & 10.73 / 10.73 (91) & 22.43 / 23.78 & \textbf{\underline{15.79}} / 11.54 (17) & 6.24 / 6.24 (30) & 27.22 / 27.22 (29) & 11.56 / 20.48 (15) & 51.36 / 53.43 (11) \\
        PCS-I+III-H & 24.36 / 24.36 (29) & \underline{11.69} / \underline{11.69} (91) & 21.27 / 23.13 & 14.69 / 9.94 (17) & 4.90 / 4.90 (30) & 28.36 / 28.36 (29) & 7.98 / 20.12 (15) & 50.40 / 52.34 (11) \\
        PCS-I+II+III-C & 24.01 / 24.01 (29) & 11.35 / 11.35 (91) & 22.74 / 24.79 & 8.55 / 6.73 (13) & 6.34 / 6.34 (29) & 31.71 / 31.71 (31) & 12.02 / 21.74 (15) & \underline{55.07} / \underline{57.41} (11) \\
        PCS-I+II+III-H & 24.30 / 24.30 (29) & 10.54 / 10.54 (91) & \underline{23.53} / \underline{25.43} &  12.94 / 11.22 (13) & 6.35 / 6.35 (29) & 32.43 / 32.43 (31) & \underline{13.37} / 22.05 (15) & 52.58 / 55.12 (11) \\
    \bottomrule
    \end{tabular}
    \end{adjustbox}
    \caption{Macro- and micro-averaged BMAcc (in percentage) as well as the predicted number of paradigm slots (in brackets), for each method. Overall best scores are \textbf{bold}, and the best scores of our system are \underline{underlined}.}
    \label{tab:results}
\end{table*}{}

\label{sec:metric}
Systems for \textit{supervised} or \textit{semi-supervised} paradigm completion are commonly being evaluated using word-level accuracy \cite{dreyer-eisner-2011-discovering,cotterell-etal-2017-conll}.
However, this is not possible for our task because our system cannot access the gold data paradigm slot descriptions and, thus, does not necessarily produce one word for each ground-truth inflected form. Furthermore, the system outputs pseudo-tags, and the mapping from pseudo-tags to paradigm slots is unknown.

Therefore, we propose to use \textit{best-match accuracy} (BMAcc), the best accuracy among all mappings from pseudo-tags to paradigm slots, for evaluation. Let $\Gamma=\{\gamma_i\}_{i=1}^{N}$ and $\hat{\Gamma}=\{\hat{\gamma_j}\}_{j=1}^{M}$ be the set of all paradigm slots in the ground truth and the prediction, respectively, with transformation functions $f:\Sigma^*\times\Gamma\rightarrow\Sigma^*\cup \{\emptyset\}$ and $\hat{f}:\Sigma^*\times\hat{\Gamma}\rightarrow\Sigma^*$,\footnote{These are equivalent to our definition in \autoref{sec:formal-description}, since the system does not need to predict the inflectional features.} where $f_\gamma(\ell)=\emptyset$ if the corresponding inflection is missing in the ground truth.\footnote{In practice, we merge paradigm slots that are identical for all lemmas in both the predictions and gold standard before evaluating.} 
We define two types of BMAcc:

\paragraph{Macro-averaged BMAcc} This is the average \textit{per-slot accuracy} for the best possible matching of slots. For any $\gamma_i,\hat{\gamma}_j$, we define $g_t(\mathcal{L},\gamma_i,\hat{\gamma}_j)$ as the number of correct guesses (true positives) if $\hat{\gamma}_j$ maps to $\gamma_i$, $g_a(\mathcal{L},\gamma_i)$ as the number of ground truth inflections for $\gamma_i$, and $\textrm{acc}(\mathcal{L},\gamma_i,\hat{\gamma}_j)$ as the per-slot accuracy:
\begin{align}
g_t(\mathcal{L},\gamma_i,\hat{\gamma}_j)&=|\{\ell\in\mathcal{L} \mid f_{\gamma_i}(\ell)=\hat{f}_{\gamma_j}(\ell)\neq \emptyset\}|\nonumber\\
g_a(\mathcal{L},\gamma_i)&=|\{\ell\in\mathcal{L} \mid {f_\gamma}_i(\ell) \neq \emptyset\}|,
\end{align}
and
\begin{align}
\textrm{acc}(\mathcal{L},\gamma_i,\hat{\gamma}_j) = \frac{g_t(\mathcal{L},\gamma_i,\hat{\gamma}_j)}{g_a(\mathcal{L},\gamma_i)}.
\end{align}

Then, we construct a complete bipartite graph with $\Gamma$ and $\hat{\Gamma}$ as two sets of vertices and $\textrm{acc}(\mathcal{L},\gamma_i,\hat{\gamma}_j)$ as edge weights. The maximum-weight full matching 
can be computed efficiently with the algorithm of \citet{karp1980algorithm}. With such a matching $\mathcal{M}=\{(\gamma_m,\hat{\gamma}_m)\}_{m=0}^{\min\{N,M\}}$, the macro-averaged BMAcc is defined as:
\begin{align}
&\textrm{BMAcc-macro}(\mathcal{L},\Gamma,\hat{\Gamma}) \\= 
&\frac{\sum_{(\gamma_m,\hat{\gamma}_m)\in\mathcal{M}}\textrm{acc}(\mathcal{L},\gamma_m,\hat{\gamma}_m)}{\max\{N,M\}} \nonumber
\end{align}
The normalizing factor $\frac{1}{\max\{N,M\}}$ rewards predicting the correct number of paradigm slots. In the case when $\textrm{acc}(\mathcal{L},\gamma_m,\hat{\gamma}_m) = 1$ for all $(\gamma_m,\hat{\gamma}_m)\in\mathcal{M}$,  $\textrm{BMAcc-macro}(\mathcal{L},\Gamma,\hat{\Gamma})$
reaches its maximum if and only if $N=M$.

\paragraph{Micro-averaged BMAcc} Our second metric
is conceptually closer to word-level accuracy. We start with the same process of bipartite graph matching, but instead use $g_t(\mathcal{L},\gamma_i,\hat{\gamma}_j)$ as edge weights. Given the optimal matching $\mathcal{M}$, the micro-averaged BMAcc is defined as:
\begin{multline}
\textrm{BMAcc-micro}(\mathcal{L},\Gamma,\hat{\Gamma})=\\
\frac{N}{\max\{N,M\}}\frac{\sum_{(\gamma_m,\hat{\gamma}_m)\in\mathcal{M}}g_t(\mathcal{L},\gamma_m,\hat{\gamma}_m)}{\sum_{\gamma_i\in\Gamma}g_a(\mathcal{L},\gamma_i)}.
\end{multline}

\subsection{Results and Discussion}

\paragraph{Overall Results } We present our experimental results in \autoref{tab:results}.
The performance of our system varies widely across languages, with best results for \lang{ENG} ($74\%$ BMAcc).
On average over languages, our final system obtains $18.76\% / 19.06\%$ BMAcc on the test set, as compared to the baseline of $4.94\% / 5.25\%$ and skylines of $18.70\% / 18.70\%$ and $35.58\% / 35.56\%$. Compared to versions of our system without selected components, our final system performs best on average for both development and test languages. 
Leaving out step II or step III leads to a reduction in performance.

Notably, variants of our system outperform the \textit{skyline} CoNLL17-1, which has seen one training example and, thus, knows the correct paradigm size in advance, on \lang{EUS}, \lang{BUL}, \lang{ENG}, \lang{FIN}, \lang{KAN}, \lang{NAV}, \lang{TUR}, \lang{MLT}, \lang{RUS}, and \lang{SWE}. Moreover, it even outperforms CoNLL17-10 on \lang{EUS}, \lang{ENG}, and \lang{NAV}, which shows that unsupervised paradigm completion has promise even in cases where a limited number of training examples---but not large amounts---are available.

\paragraph{Differences between Languages } We hypothesize that the large differences between languages---over $73\%$ between \lang{EUS} and \lang{ENG}---can be explained in parts by the following reasons:
    
    Intuitively, the larger the paradigm, the more difficult the task. If the number of slots is huge, each individual inflection is rare, and it is hard for any unsupervised paradigm completion system to distinguish true inflections (e.g., \word{rise} $\to$ \word{rises}) from false candidates (e.g., \word{rise} $\to$ \word{arise}). This could explain the high performance on \lang{ENG} and the low performance on \lang{EUS}, \lang{FIN}, and \lang{FAS}.
    
    Related to the last point, in a limited corpus such as the Bible, some inflected forms might not appear for any lemma, which makes them undetectable for unsupervised paradigm completion systems. For example, 
    a \lang{FAS} paradigm has 136 slots in UniMorph, but only 46 are observed.\footnote{We had to estimate that number based on available data in UniMorph: a slot is considered observed if at least one inflected form for that slot can be found.} Additional statistics can be found in \autoref{tab:ana}.
    
    Furthermore, Assumption 2 does not hold for all languages. Surface forms can be shared between paradigm slots, as, for instance, in English for \textit{he \textbf{studied}} and \textit{he has \textbf{studied}}.
    Different languages show different degrees of this phenomenon called \textit{syncretism}.

\paragraph{Pipeline Effects } Different combinations of components result in major performance differences. In particular, each step of our system has the potential to introduce errors. This demonstrates a pitfall of pipeline methods also discussed in \citet{mccarthy-etal-2020-skinaugment}: the quality of individual steps, here, e.g., \textsc{edit tree} discovery and retrieval of additional lemmas, can greatly affect the results of PCS-II and PCS-III.

\paragraph{Differences in Components } Details of individual components also affect the results. On the one hand, applying more than one iteration of additional lemma retrieval impacts the results only slightly, as those lemmas are assigned very small weights. On the other hand, we see performance differences $>2\%$ between PCS-III-C and PCS-III-H for \lang{DEU}, \lang{MLT}, and \lang{SWE}.

\begin{table}[tbp]
    \centering
    \small
    \begin{tabular}{cccc}
        \toprule
            \textbf{Language} & \textbf{\shortstack{\textsc{ET} \\ Match}} & \textbf{\shortstack{Rep. \\ Words}} & \textbf{\shortstack{Absent/Total \\ Slots}} \\
        \midrule
           \lang{MLT} & 18.13 & 58.18 & 0/16\\
           \lang{FAS} & 12.10 & 28.66 & 90/136\\
           \lang{POR} & 56.52 & 30.56 & 4/76\\
           \lang{RUS} & 24.69 & 25.45 & 4/55\\
           \lang{SWE} & 72.91 & 13.98 & 0/11\\
        \bottomrule
    \end{tabular}
    \caption{Statistics for our development languages, computed with UniMorph. \textit{ET Match} is the percentage of gold $(\ell, w)$ pairs that can be matched to an \textsc{edit tree}. \textit{Rep.~Words} denotes the percentage of inflected forms that represent multiple paradigm slots.
    \textit{Absent/Total Slots} is the numbers of unobservable and total slots.}
    \label{tab:ana}
\end{table}

\paragraph{Analysis of \textsc{Edit Tree} Quality }
As it is the first step in our pipeline, the quality of the \textsc{edit tree} discovery strongly affects the performance of later components. For our development languages, we show in \autoref{tab:ana} the percentage of $(\ell, w)$ pairs for which the system predicts an \textsc{edit tree} $\psi$ such that $\psi(\ell) = w$ appears in the gold paradigm of~$\ell$. This corresponds to the highest possible performance after PCS-I. \lang{FAS} has the worst performance ($12.10\%$), while the results for \lang{SWE} are high ($72.91\%$). As expected, languages with lower values here 
also obtain lower final results.
 
\paragraph{Analysis of Syncretism }
We further hypothesize that syncretism could be a  source of errors, due to \autoref{thm:oneslotpertree}. \autoref{tab:ana} shows the percentage of words that are the inflected forms corresponding to multiple paradigm slots of the same lemma.

We observe that \lang{SWE} has a low degree of syncretism, and, in fact, our system predicts the correct paradigm size for \lang{SWE}. A high degree of syncretism, in contrast, might contribute to the low performance on \lang{MLT}. 

\section{Conclusion}
We proposed \textit{unsupervised morphological paradigm completion}, a novel morphological generation task. We further developed a system for the task, which performs the following steps: (i) \textsc{edit tree} retrieval, (ii) additional lemma retrieval, (iii) paradigm size discovery, and (iv) inflection generation. Introducing best-match accuracy, a metric for the task, we evaluated our system 
on a typologically diverse set of 14 languages. 
Our system obtained promising results for most of our languages and even outperformed a minimally supervised baseline on Basque, English, and Navajo.
Further analysis showed the importance of our individual components and detected possible sources of errors, like wrongly identified \textsc{edit trees} early in the pipeline or syncretism.

In the future, we will explore the following directions: (i) A difficult challenge for our proposed system is to correctly determine the paradigm size.
Since transfer across related languages has shown to be beneficial for morphological tasks \citep[\emph{inter alia}]{jin-kann-2017-exploring,mccarthy-etal-2019-sigmorphon, anastasopoulos-neubig-2019-pushing}, future work could use typologically aware priors to guide the number of paradigm slots based on the relationships between languages. 
(ii) We plan to explore other methods, like word embeddings, to incorporate context information into our feature function. (iii) We aim at developing better performing string transduction models for the morphological inflection step. By substituting the current transducers in our pipeline, we expect that we will be able to improve the overall performance of our system. 

\section*{Acknowledgments}
We are grateful to the group of SIGMORPHON 2020 shared task organizers as well as Alexander Erdmann for discussions about our proposed task in summer 2019. Special thanks go to Mans Hulden, who was the first to suggest an initial version of the task, and to Jason Eisner, the president of SIGMORPHON. Last but not least, we also thank Clara Vania and the anonymous reviewers for their helpful comments on this work.

\bibliography{acl2020,anthology}
\bibliographystyle{acl_natbib}

\end{document}